\documentclass[11pt]{article}

\usepackage[english]{babel}

\usepackage{graphicx}
\usepackage[utf8]{inputenc}
\usepackage[
      backend=biber,          
      style=authoryear,       
      natbib,
      hyperref=true
    ]{biblatex}
\usepackage[T1]{fontenc}
\usepackage{hyperref}
\hypersetup{
    colorlinks=true, 
    linkcolor=blue,    
    citecolor=blue,    
    urlcolor=blue      
}
\usepackage[letterpaper,top=2cm,bottom=2cm,left=3cm,right=3cm,marginparwidth=1.75cm]{geometry}

\usepackage{amsmath}
\usepackage{graphicx}

\title{A Survey on Collaborative Mechanisms Between Large and Small Language Models}
\author{Yi Chen\textsuperscript{1,2}, \ Haiquan Wang\textsuperscript{1,2}, \ JiaHao Zhao\textsuperscript{1,2}, \ HaoHao Han\textsuperscript{1,2} \\
     \textsuperscript{1}Chengdu Institute of Computer Applications, Chinese Academy of Sciences, Chengdu, China \\
     \textsuperscript{2}University of Chinese Academy of Sciences, Beijing, China \\
     \texttt{\{chenyi24,wanghaiquan22,zhaojiahao241,hanhaohao24\}@mails.ucas.ac.cn} \\
}
\date{}

\DeclareCiteCommand{\parencite}[\mkbibparens]
  {\usebibmacro{prenote}}
  {\usebibmacro{citeindex}%
   \printtext[bibhyperref]{%
     \usebibmacro{cite}}}
  {\multicitedelim}
  {\usebibmacro{postnote}}
  
\begin{document}

\maketitle

\begin{center}
    \section*{Abstract} 
\end{center}
Large Language Models (LLMs) deliver powerful AI capabilities but face deployment challenges due to high resource costs and latency, whereas Small Language Models (SLMs) offer efficiency and deployability at the cost of reduced performance. Collaboration between LLMs and SLMs emerges as a crucial paradigm to synergistically balance these trade-offs, enabling advanced AI applications, especially on resource-constrained edge devices. This survey provides a comprehensive overview of LLM-SLM collaboration, detailing various interaction mechanisms (pipeline, routing, auxiliary, distillation, fusion), key enabling technologies, and diverse application scenarios driven by on-device needs like low latency, privacy, personalization, and offline operation. While highlighting the significant potential for creating more efficient, adaptable, and accessible AI, we also discuss persistent challenges including system overhead, inter-model consistency, robust task allocation, evaluation complexity, and security/privacy concerns. Future directions point towards more intelligent adaptive frameworks, deeper model fusion, and expansion into multimodal and embodied AI, positioning LLM-SLM collaboration as a key driver for the next generation of practical and ubiquitous artificial intelligence.
\vspace{2em}

\section{Introduction}

\subsection{Research Background and Motivation}
In recent years, Large Language Models (LLMs) have achieved breakthrough progress in fields such as natural language processing, code generation, and intelligent question answering \citep{brown2020language, openai2023gpt4}. However, as the scale of model parameters continues to grow, their consumption of computing resources, energy costs, and deployment expenses have also risen significantly. Particularly in edge scenarios requiring low latency and high privacy, such as smartphones, IoT devices, and edge servers, the traditional cloud-based LLM inference model faces serious feasibility challenges \citep{zhou2024benchmark}. Concurrently, Small Language Models (SLMs) are widely used in resource-constrained devices due to their lightweight structure, fast inference speed, and ease of deployment \citep{gao2025strategic}. Against this backdrop, the collaborative mechanism between large and small language models has gradually become an important research direction in both industry and academia. It aims to leverage the powerful capabilities of LLMs and the high efficiency of SLMs through intelligent collaboration, building a more intelligent, efficient, and reliable inference system.

\subsection{Definition and Scope of Large-Small Model Collaboration}
Broadly speaking, large-small model collaboration refers to a mechanism where LLMs and SLMs work together within a system, complementing each other's strengths. This paradigm can be divided into several sub-directions, such as pipeline collaboration, parallel collaboration, conditional triggered inference, and knowledge distillation \citep{wang2024comprehensive, gao2025strategic}. Pipeline collaboration is a sequential execution mode where the output of one model serves as the input for another \citep{wang2024comprehensive}. Since SLMs typically offer higher efficiency, they are often used for preliminary processing or generating candidate results, which are then passed to LLMs for more complex reasoning or knowledge integration \citep{gao2025strategic}. In the Cascade Speculative Decoding framework, an SLM generates draft responses, while an LLM performs verification and correction in parallel, thereby improving response speed \citep{chen2023speculative}. In terms of conditional inference, research has proposed using confidence scores to determine whether to invoke an LLM, thus achieving on-demand activation \citep{gupta2023adaptive}. Furthermore, knowledge distillation allows the knowledge from an LLM to be compressed and transferred to an SLM through training, enhancing the latter's ability to model complex tasks while retaining its computational advantages \citep{gu2024minillm}. These studies constitute the core scope of large-small model collaboration and provide a theoretical and practical foundation for building energy-efficient AI systems.

\subsection{The Rise of On-Device Large Models and Its Impetus for Collaboration Research}
The emergence of on-device large language models is propelling research into collaborative mechanisms to a new stage. In recent years, with improvements in chip performance and the maturation of model compression techniques, several tech giants have begun deploying proprietary large models on terminals. In 2024, Apple integrated an on-device language model with approximately 3 billion parameters into its "Apple Intelligence" system to handle some natural language tasks locally, while complex requests are delegated to cloud-based models \citep{apple2024ai}. Similarly, Huawei introduced a lightweight version of its Pangu large model into its HarmonyOS intelligent assistant, enabling hybrid edge-cloud inference \citep{huawei2024pangu}. This architecture prompts researchers to delve deeper into questions such as: How can the task boundaries between LLMs and SLMs be allocated more effectively? How can computation offloading and dynamic routing be managed between the edge and the cloud? How can collaboration mechanisms improve overall energy efficiency and response quality? Related challenges include model selection strategies, edge-cloud communication overhead, privacy protection mechanisms, and multi-model fusion accuracy \citep{zhang2025efficient}. Therefore, studying the collaboration mechanisms of large and small models is not only a hot topic for theoretical exploration but also a key technological path for intelligent terminals moving towards localized, high-performance, and low-power AI inference.

\subsection{Structure and Contributions of This Paper}
This paper first provides an introduction, outlining the research background, motivation, and the definition and scope of large-small model collaboration. Next, it elaborates on related concepts and foundations, including the characteristics, advantages, and limitations of large and small language models, as well as the basis and necessity for their collaboration. The third section delves into the mechanisms and architectures of large-small model collaboration, classifying and introducing modes such as pipeline, hybrid/routing, auxiliary/enhancement, knowledge distillation-driven, and integration/fusion, and analyzes the key technologies for implementing collaboration. Subsequently, driven by on-device requirements, it discusses the application scenarios of collaborative mechanisms, categorized by real-time low-latency inference, privacy sensitivity, task-specific customization, offline/weak network environments, and energy constraints. Finally, it presents current challenges and open questions, looks ahead to future development trends, and summarizes the research significance, value, and reflections of the entire paper.


\section{Related Concepts and Foundations}

Before delving into the collaborative mechanisms between Large Language Models (LLMs) and Small Language Models (SLMs), it is necessary to systematically review the basic concepts, structural characteristics, advantages, and limitations of these two types of models. They exhibit significant differences in model capabilities, resource consumption, deployment scenarios, and research paths. It is precisely this heterogeneity that forms the fundamental motivation for research on model collaboration.

\subsection{Large Language Models}

\subsubsection{Definition, Architecture, and Characteristics}

Large Language Models (LLMs) typically refer to language understanding and generation models with parameter scales in the billions or more. Their architectures are mostly based on the Transformer framework, employing autoregressive (e.g., GPT series) or masked language modeling strategies (e.g., BERT, T5) for pre-training \citep{brown2020language, raffel2020exploring, openai2023gpt4}. These models learn language structure, knowledge representation, and reasoning patterns from massive unsupervised corpora, enabling deep modeling of natural language text and supporting generalization to various downstream tasks.

From an architectural perspective, LLMs usually consist of tens to hundreds of Transformer layers, each containing multi-head attention mechanisms, feed-forward networks, residual connections, and normalization operations. As the number of model parameters increases, their capabilities do not grow linearly but exhibit certain "emergent phenomena." That is, once the model scale exceeds a certain threshold, they automatically demonstrate more complex abilities such as language understanding, mathematical reasoning, and multimodal interaction \citep{wei2022emergent, gao2025strategic}. This non-linear performance leap has become an important clue in the current exploration of artificial general intelligence.

\subsubsection{Advantages and Limitations}

LLMs possess unparalleled advantages in general language modeling. They exhibit strong generalization capabilities and can perform various natural language processing tasks through prompts or in-context learning, including question answering, translation, summarization, code generation, and knowledge retrieval, demonstrating potential for zero-shot and few-shot learning. LLMs have deep semantic understanding and long-text modeling capabilities, preserving contextual logic and structural integrity. Through fine-tuning or reinforcement learning (such as RLHF), their behavior and alignment with human preferences can be further optimized.

However, the limitations of LLMs cannot be ignored. Firstly, resource consumption is extremely high. Training often requires high-performance computing clusters (such as GPUs/TPUs) and weeks to months of time cost, while the inference stage also faces issues like large memory footprint and high latency. Their large size makes them difficult to deploy on resource-constrained terminal or edge devices, limiting their real-time application scenarios. Due to the uncontrollable nature of data during pre-training, models may exhibit factual errors, bias amplification, and hallucination risks, increasing the safety and ethical costs in practical applications \citep{ zhang2025efficient}. The closed-source deployment of LLMs and the reliance on cloud platforms for large model calls also raise concerns about data privacy and compliance \citep{zhou2024benchmark}.

\subsection{Small Language Models}

\subsubsection{Definition and Characteristics}

Small Language Models (SLMs) refer to models with parameter scales ranging from millions to a few hundred million. Their design goal is to balance language modeling capabilities with deployment efficiency, significantly reducing computational overhead and hardware requirements while ensuring basic language understanding capabilities. Structurally, SLMs are often compressed versions of large models, employing fewer Transformer layers, smaller hidden dimensions, and fewer attention heads, while streamlining the computation flow in the inference path as much as possible \citep{gu2024minillm, apple2024ai}.

Compared to LLMs, SLMs are more suitable for deployment on platforms with limited computing resources, such as edge devices, mobile terminals, and browser plugins. Their characteristics, including fast inference speed, low energy consumption, and flexible deployment, make them key components for achieving localized, low-latency human-computer interaction. In application scenarios like voice assistants, text input methods, and intelligent customer service, small models can respond quickly to user requests, protect user privacy, and reduce reliance on networks and the cloud.

\subsubsection{Advantages, Limitations, and Typical Construction Methods}

The prominent advantages of small models are reflected in their lightweight nature, high efficiency, and deployment-friendliness. On the one hand, due to their compact structure, they can perform inference on ordinary CPUs or low-power devices without relying on high-performance GPUs, significantly reducing operational costs. On the other hand, SLMs typically possess good task adaptability and can achieve performance optimization in specific scenarios through fine-tuning or distillation. Furthermore, keeping data processing local helps enhance user privacy protection capabilities, meeting the requirements of certain industries with high compliance demands.

However, it is undeniable that SLMs still have significant gaps compared to LLMs in terms of knowledge scale, semantic expression, and reasoning depth. Due to limited model capacity, they struggle to cover knowledge across a wide range of domains and often exhibit insufficient understanding, vague generation, or logical incoherence when facing open-domain tasks. In handling tasks such as long text processing, complex reasoning, and multi-turn dialogues, small models often find it difficult to maintain semantic coherence and information consistency. Additionally, their generalization ability is weaker, performing less well than LLMs in cross-domain transfer or zero-shot learning \citep{gupta2023adaptive}.

Currently, the mainstream construction methods for SLMs can be categorized into three types: First, knowledge distillation, which compresses the capabilities of large models into small models using a teacher-student framework, such as DistilGPT and TinyLLaMA; second, building lightweight models from scratch through independent training on medium-scale corpora, such as MiniBERT and ALBERT; third, pruning, quantization, or low-rank decomposition based on existing model architectures to compress size while retaining some original performance \citep{wang2024comprehensive}. These methods collectively promote the widespread application of small models in industrial deployment and provide a technical reserve for subsequent research on collaborative mechanisms.

\subsection{Foundation and Necessity of Collaboration Between Large and Small Language Models}
\subsubsection{Technical Foundation for Collaboration: Complementarity of Capabilities and Characteristics}  
Large Language Models (LLMs) and Small Language Models (SLMs) exhibit significant differences in capabilities, architecture, and deployment characteristics, forming a natural technical foundation for collaboration.  

\subsubsection{Complementarity of Capability Boundaries}  
The advantages of large models include strong contextual understanding, logical reasoning, and multi-task generalization capabilities (e.g., GPT-4 achieves over 90\% accuracy on complex tasks like mathematical reasoning and code generation), but they rely on massive training data (trillion-level parameters, hundred-billion-level corpora) and high computational power support (single inference cost is about $0.1 - $1). 

Conversely, the advantages of small models lie in their smaller parameter count (typically under 1 billion parameters, e.g., TinyBERT, DistilBERT), flexible deployment (can run on edge devices like phones, smart speakers), and fast inference speed (latency below 10ms, 1/10th that of large models). However, their capabilities are limited in complex logic and long-context processing (e.g., accuracy on math problems is less than 50\%).

Collaboration essentially involves the large model acting as the "brain" for complex decision-making, while the small model serves as the "nerve endings" for lightweight interaction, forming a "central processor + edge node" division of labor system.

\subsubsection{Transferability of Architecture and Training}  
This can be achieved through points like knowledge distillation techniques and Parameter-Efficient Fine-Tuning (PEFT). Knowledge distillation techniques, using methods like Soft Labels and intermediate layer feature transfer (e.g., FitNets), allow large models to transfer implicit knowledge (such as semantic representation, reasoning logic) into training signals for small models. This enables small models to inherit the core capabilities of large models while remaining lightweight (e.g., TinyBERT achieves 96\% of BERT's performance on the GLUE benchmark with 75\% fewer parameters). 

Parameter-Efficient Fine-Tuning, through techniques like LoRA and QLoRA, allows small models to perform localized fine-tuning based on the pre-trained parameters of large models, quickly adapting to specific tasks (e.g., in medical Q\&A scenarios, the accuracy of small models increased by 20\% after fine-tuning). This forms a collaborative training paradigm of "large model foundation + small model specialization." 

\subsubsection{Adaptability of Deployment Scenarios}  
In terms of adaptability, collaboration between cloud and edge can be utilized. Large models rely on high-compute cloud servers (requiring tens of GB of VRAM) and are suitable for handling batch complex tasks (e.g., text generation, multi-turn dialogue). Small models can be deployed on low-compute devices (e.g., Raspberry Pi, mobile phone chips) and are responsible for real-time interaction and local data preprocessing (e.g., speech recognition, preliminary user intent analysis). The two communicate via APIs or lightweight protocols (e.g., gRPC) to enable data flow. 

\subsubsection{Necessity of Collaboration: An Inevitable Choice Driven by Real-world Needs}  
In the collaboration process between large and small models, addressing real-world needs is essential. Issues to consider include computational cost constraints, real-time requirements, handling of certain hierarchical tasks, and multimodality, among others.

Specifically, regarding computational power, a single inference by a large model consumes hundreds of times more power than a small model (e.g., GPT-4 generating 1000 words consumes about 500 times the energy of TinyBERT). In high-frequency interaction scenarios (e.g., customer service, real-time translation), relying solely on large models would lead to soaring costs (enterprise-level application monthly compute costs could exceed millions of yuan). 
Furthermore, edge devices (e.g., autonomous driving systems) require response latency below 50ms. Cloud-based inference latency for large models is typically over 500ms, failing to meet the demand. Local deployment of small models can achieve sub-millisecond responses. Collaboration with large models can compress end-to-end latency to within 200ms (e.g., voice interaction systems in smart cars). Complex tasks (e.g., legal document generation, scientific paper polishing) are handled by large models, while simple tasks (e.g., keyword extraction, sentiment classification) are processed by small models, forming a "pyramid-style" task allocation system. In e-commerce customer service, a small model first identifies user intent (90\% accuracy); if it's a complex query (e.g., interpreting return policies), it forwards it to the large model, increasing efficiency by 40\% compared to a single-model approach.  

Small models can preprocess data from modalities like images and speech (e.g., speech-to-text, image feature extraction). Large models perform cross-modal reasoning (e.g., generating personalized responses based on user voice and expression), reducing the input complexity for large models (e.g., multimodal processing latency reduced by 30\%).  
Fields like healthcare and finance require user data to remain local. Small models can perform data cleaning and anonymization on the device (e.g., removing names and IDs from medical records), transmitting only anonymized features to the large model, thus avoiding privacy leakage risks (e.g., in federated learning, small models train parameters locally, while the large model aggregates global knowledge). The simple structure of small models (e.g., single-layer Transformer or CNN) allows their decision logic to be explained using visualization tools (e.g., attention heatmaps), compensating for the "black box" nature of large models (e.g., in medical diagnosis scenarios, the small model provides a preliminary conclusion, which the large model verifies and generates an explanatory text).  

Beyond these issues, application scenarios need consideration. Large models struggle to adapt to niche domains (e.g., dialect recognition, vertical industry knowledge bases). Small models can be customized for specific scenarios (e.g., a small model for the tax domain with only 1 billion parameters achieves 95\% accuracy in policy interpretation), forming a collaborative ecosystem of "general foundation + domain plugin" with large models. Small and medium-sized enterprises and developers often cannot afford the computational power and training costs of large models. By combining small models with cloud-based large model APIs (e.g., using OpenAI's GPT-4 API with a local small model for pre-filtering), deployment costs can be reduced by over 90\%, promoting the democratization of AI technology.  

SLMs often handle the front-end tasks in a model pipeline, such as input understanding, candidate generation, or decision-making, making them a crucial part of achieving efficient collaboration \citep{gao2025strategic}. In complex systems, SLMs handle front-end basic tasks (like input understanding, candidate generation), passing the preliminary processed results to large models (LLMs) for deep reasoning, forming a collaboration mode of "lightweight model preprocessing + heavyweight model refinement." In intelligent customer service systems: An SLM quickly analyzes the user's question ("check order status"), extracts key information (order number), and generates candidate query solutions. The LLM then uses these candidate solutions to call database interfaces and generate the final response.
SLMs (Small Language Models), with their efficiency and specificity, have become indispensable "front-end processors" in AI systems. Through division of labor with LLMs, they significantly improve overall performance and resource utilization. This architecture is widely applied in industry and is an important practice for realizing the "Large Model + Small Model" hybrid ecosystem.

\section{Collaboration Mechanisms and Architectures for Large and Small Models}

\subsection{Classification of Collaboration Modes}
Based on the interaction methods and information flow between LLMs and SLMs, their collaboration modes can be classified into pipeline, hybrid/routing, auxiliary/enhancement, knowledge distillation-driven, and integration/fusion collaboration.
\subsubsection{Pipeline Collaboration}
Pipeline collaboration is a sequential execution mode where the output of one model serves as the input for another \citep{wang2024comprehensive}. Since SLMs typically offer higher efficiency, they are often used for executing preliminary processing or generating candidate results, which are then passed to LLMs for more complex reasoning or knowledge integration \citep{gao2025strategic}.

In recommendation systems, pipeline collaboration is widely used. An LLM can generate a series of candidate recommendation items based on user historical behavior and preferences, leveraging the LLM's strong ability to capture user preferences \citep{lv2025collaboration}. Then, an SLM deployed on the local device can re-rank these candidate items based on the user's real-time interaction behavior, thereby more accurately reflecting the user's current interests. This approach utilizes the LLM's global understanding capabilities while also accommodating the SLM's rapid response to real-time data \citep{lv2025collaboration}.

\begin{figure}[t]
 \centering
  \includegraphics[width=0.6\columnwidth]{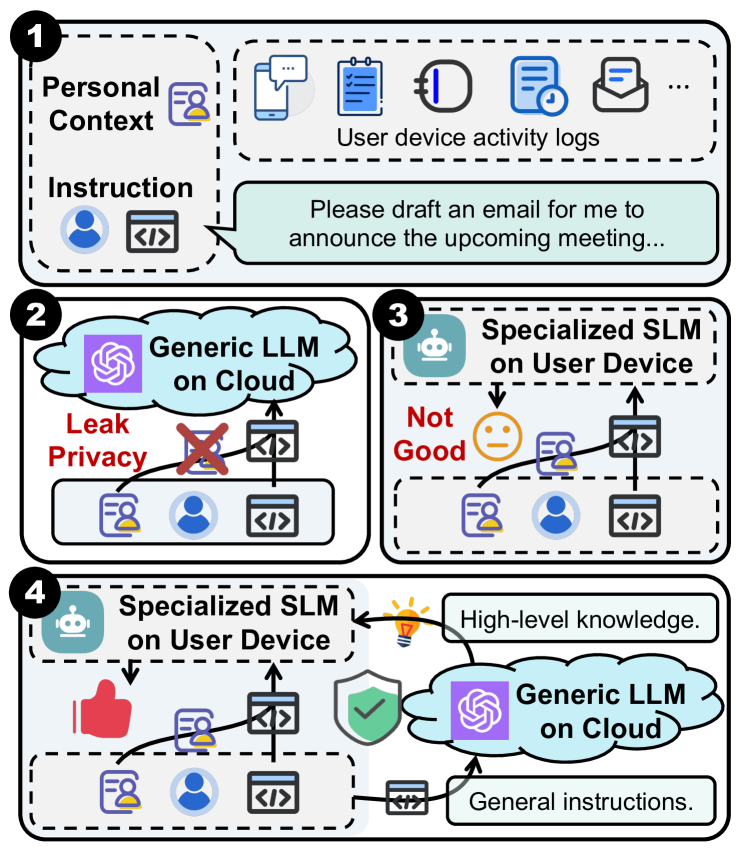} 
  \caption{CoGenesis framework structure diagram. 1. Context-aware instructional examples. 2. Context-aware Language Models (LLMs) excel in context awareness but pose privacy risks. 3. On-device specialized Small Language Models (SLMs) prioritize privacy but have lower performance. 4. Collaborative LLMs and SLMs enhance privacy protection and improve performance.}
  \label{fig:CoGenesis}
\end{figure}

The CoGenesis framework is another typical example of pipeline collaboration, as shown in Figure~\ref{fig:CoGenesis}. In this framework, an SLM deployed on the user's local device can access the user's private data and activity logs and process instructions based on this information. For tasks requiring deeper reasoning, the SLM's output can be passed as input to an LLM deployed in cloud infrastructure. This design protects user privacy while leveraging the powerful capabilities of large models. Additionally, SLMs can be used to extract key information from input text or generate concise prompts, which are then sent to LLMs to guide them in generating more relevant outputs \citep{zhang2024cogenesis,wang2024comprehensive}.

The key to pipeline collaboration lies in reasonably dividing tasks and designing effective inter-model interfaces to ensure that information can be accurately and efficiently transferred between different models \citep{wang2024comprehensive,liu2025towards,chen2025knowledge,xu2024slmrec}. The effectiveness of this mode largely depends on the ability of the first model to extract relevant information and the quality of the information passed to the second model. If the SLM fails to accurately capture key information or passes information in an inappropriate format, the LLM's performance will be adversely affected. Therefore, communication protocols and the type of information exchanged must be carefully considered to ensure seamless and effective collaboration. Pipeline collaboration typically leverages the efficiency of SLMs for preliminary processing or context collection, offloading computationally intensive and knowledge-dependent tasks to LLMs, thereby striking a balance between speed and accuracy.

\subsubsection{Hybrid/Routing Collaboration}
Hybrid or routing collaboration refers to using a mechanism (usually called a router) to decide which model (LLM or SLM) should handle a specific input or sub-task \citep{wang2025mixllm,zheng2025citer,lv2025collaboration}. The routing decision is typically based on task complexity, domain, cost, required latency, or other predefined criteria.

\begin{figure}[t]
 \centering
  \includegraphics[width=0.9\columnwidth]{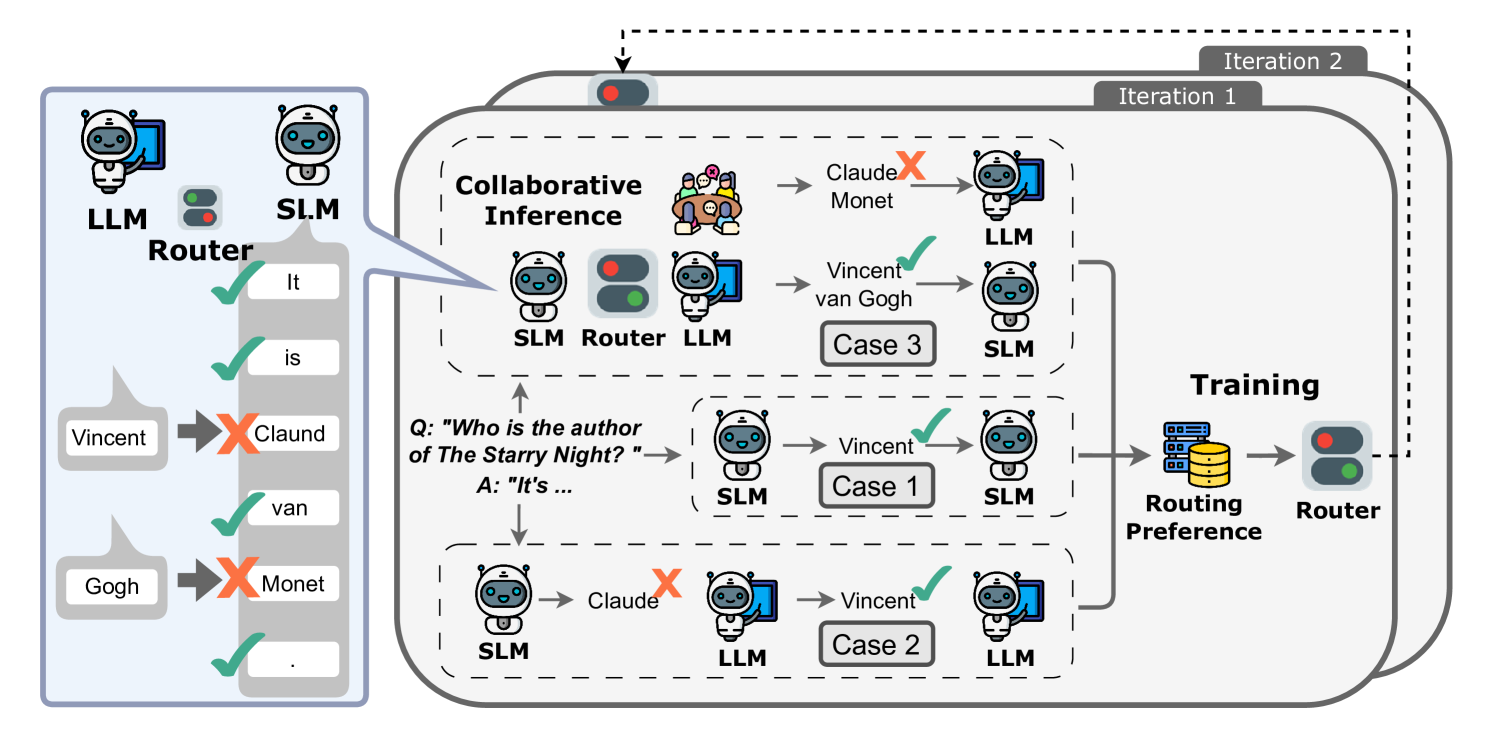} 
  \caption{CITER framework structure diagram. Utilizes a router for collaborative inference between SLM and LLM. The router is trained using routing preferences collected through three scenarios. Scenario 1: SLM generates the correct token, routing preference assigned to SLM. Scenario 2: SLM generates an incorrect token, while LLM generates the correct token, routing preference assigned to LLM. Scenario 3: Neither SLM nor LLM generates the correct token; collaborative reasoning is performed to obtain the complete response for assigning routing preference.}
  \label{fig:CITER}
\end{figure}

The CITER framework is a typical example of routing collaboration, shown in Figure~\ref{fig:CITER}. This framework adopts a token-level routing strategy, routing non-critical tokens to the SLM to improve efficiency, while routing critical tokens to the LLM to ensure generation quality. This fine-grained routing allows the system to dynamically adjust between efficiency and quality \citep{zheng2025citer}.

Other research efforts focus on developing routers capable of selecting the most appropriate LLM from a pool of candidate models \citep{varangot2025doing,wang2025mixllm}. This approach can be extended to include SLMs, enabling the router to choose among models of different scales and capabilities based on task requirements. HybridLLM utilizes a binary classifier to predict query difficulty and routes between different models accordingly \citep{yao2025toward}.

Cascading is a common routing strategy that first attempts to process input using smaller LLMs and passes the query to larger LLMs if necessary \citep{chen2025harnessing,ong2024routellm}. This method can reduce computational costs while ensuring performance.

Intelligent routing requires a mechanism capable of accurately assessing input features and the capabilities of available models to make optimal decisions. The router needs to understand the complexity of the task, the domain of the query, and the specific strengths and weaknesses of each LLM and SLM in the system. This might involve training a separate model to predict performance or using heuristic methods based on query features. Balancing cost, latency, and quality is a key consideration when designing routing mechanisms. More powerful models typically come with higher costs and latency, so the routing strategy must weigh these factors according to the application's needs. For real-time applications, latency might be the most critical factor, potentially prioritizing faster (but possibly less accurate) SLMs for certain query types. For tasks requiring high accuracy, the system might route to a more powerful (and more expensive) LLM. Routing strategies need to be configurable to meet different performance objectives.
\subsubsection{Auxiliary/Enhancement Collaboration}

In auxiliary or enhancement collaboration, one model (either LLM or SLM) assists another to improve its performance or capabilities \citep{wang2024comprehensive,shao2025division,hu2024auxiliary}.

An LLM can decompose a complex query into several sub-problems and then assign these sub-problems to an SLM for processing, or vice versa \citep{shao2025division,xu2025collab}. SLMs can provide LLMs with contextual information or domain-specific knowledge to enhance the LLMs' reasoning capabilities. LLMs can also be used to generate training data or provide feedback signals for SLMs, thereby helping SLMs learn and improve \citep{deng2023mutual}.

\begin{figure}[t]
 \centering
  \includegraphics[width=0.8\columnwidth]{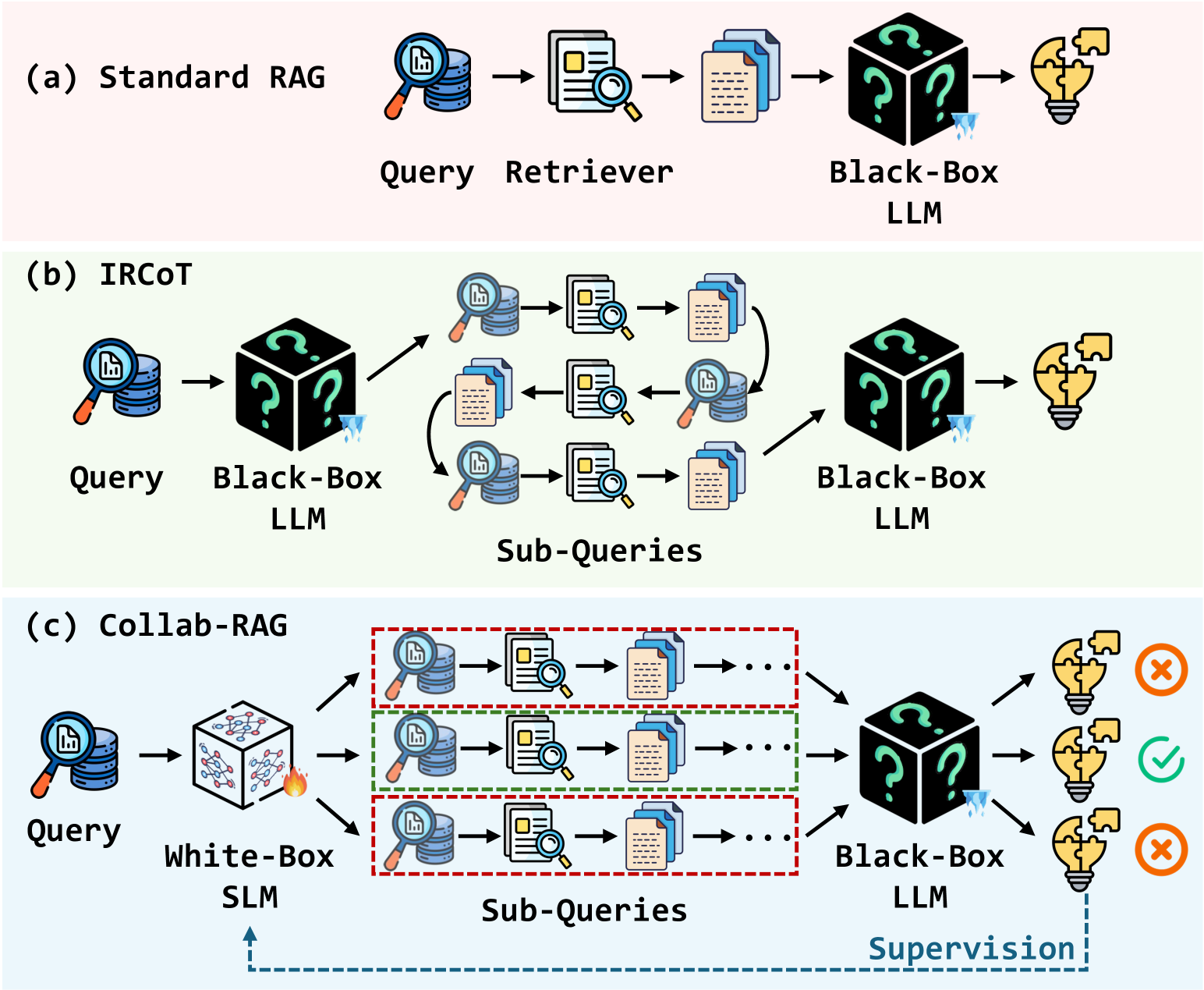} 
  \caption{Iterative training framework of Collab-RAG. The SLM updates its parameters based on the generation quality of the LLM reader. This process iterates multiple times, progressively improving the SLM's decomposition capability.}
  \label{fig:Collab-RAG}
\end{figure}

The Collab-RAG framework demonstrates an application of auxiliary collaboration, as shown in Figure~\ref{fig:Collab-RAG}. In this framework, an SLM is responsible for decomposing user queries into simpler sub-problems to facilitate the retrieval of relevant information from a knowledge base, thereby enhancing the LLM's reasoning ability when answering complex questions \citep{xu2025collab}.

Auxiliary collaboration allows for a division of labor based on the inherent strengths of each model type, potentially improving overall performance compared to using either model alone. LLMs excel at understanding complex instructions and possess broad knowledge, while SLMs may be more specialized and efficient on certain specific tasks. By having them assist each other, the system can leverage the LLM's understanding to guide the SLM, or use the SLM's specialized knowledge to inform the LLM's reasoning. Designing effective auxiliary mechanisms requires careful consideration of how the models can best complement each other and the communication methods used for this interaction. Simply having one model feed its output to another might not be the most effective approach. Assistance might involve providing specific types of information, guiding the reasoning process, or offering feedback on the other model's performance. Communication needs to be tailored to the specific assistance being provided.
\subsubsection{Knowledge Distillation-Driven Collaboration}
Knowledge Distillation (KD) is a key technique used to transfer knowledge and capabilities from a large, often proprietary teacher model (LLM) to a smaller, more efficient student model (SLM) \citep{xu2024survey,xu2024slmrec}.

KD involves multiple aspects, including the type of knowledge (output probabilities, intermediate representations), distillation algorithms (supervised fine-tuning, divergence minimization), and application domains (model compression, skill transfer, domain specialization). Data augmentation also plays an important role in enhancing the effectiveness of KD for LLMs. Depending on whether the teacher model can provide internal information, KD methods can be divided into white-box KD and black-box KD. KD is also used to transfer specific skills, such as reasoning, instruction following, and tool use \citep{xu2024survey,hendriks2025honey,gu2023minillm}.

Knowledge distillation is a critical mechanism for enabling LLM-level intelligence to be deployed in resource-constrained environments by creating smaller, more efficient models. While LLMs offer superior performance, their size and computational requirements limit their applicability in many practical scenarios. KD allows us to transfer the knowledge learned by these large models into smaller models that can be deployed on edge devices or used in applications with strict latency requirements. The success of knowledge distillation depends on several factors, including the choice of teacher and student models, the quality and quantity of distillation data, and the specific distillation techniques employed. Different LLMs have different strengths, and the choice of teacher model will influence the type of knowledge transferred. The distillation data should be representative of the tasks the student model will perform. Selecting the appropriate distillation algorithm and tuning its parameters are also crucial for achieving optimal performance.

\subsubsection{Integration/Fusion Collaboration}
Integration or fusion collaboration refers to combining the architectures or outputs of LLMs and SLMs into a unified system to leverage their complementary strengths in a more tightly coupled manner \citep{naveed2023comprehensive,subramanian2025small,lv2025collaboration,dong2024hymba,wan2024knowledge}.

\begin{figure}[t]
 \centering
  \includegraphics[width=0.8\columnwidth]{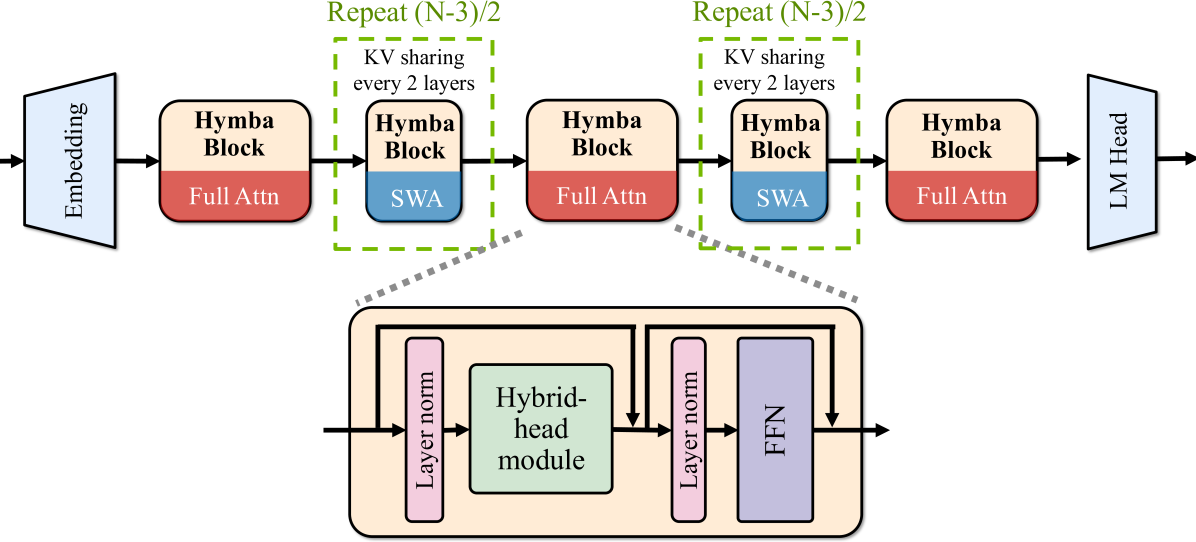} 
  \caption{Overall architecture of the Hymba model}
  \label{fig:Hymba}
\end{figure}

Hymba is a hybrid architecture, shown in Figure~\ref{fig:Hymba}, which integrates the attention mechanism of transformers and state-space models (SSMs) within the same layer, thereby achieving high recall and efficient context summarization \citep{dong2024hymba}. Model fusion techniques refer to combining the parameters or predictions of multiple models. This can include weight averaging, ensemble methods, or more complex fusion architectures. Multimodal LLMs (MLLMs) are another example of integration collaboration, typically using an LLM as a backbone and integrating information from different modalities (such as text and images).

Integration collaboration offers the potential to create more powerful and versatile AI systems by deeply combining the strengths of different model types at the architectural level. Model fusion provides a cost-effective way to create more powerful models without extensive retraining by combining the knowledge and capabilities of existing pre-trained models. However, effectively fusing models with different architectures and ensuring the resulting model retains desired properties remain challenging.

\subsection{Key Technologies for Implementing Collaboration}
To effectively implement collaboration between LLMs and SLMs, the following key technologies are required:
\subsubsection{Task Allocation and Intelligent Routing}

In heterogeneous LLM-SLM systems, task allocation faces numerous challenges, requiring consideration of task complexity, required knowledge, computational resources, latency constraints, and cost \citep{varangot2025doing}. The main goal is to intelligently assign system input tasks to the most appropriate model for processing.

Through dynamic complexity-aware routing techniques, the system can decide which model to use based on task characteristics (such as task type, context length, domain specificity) and input complexity. In systems like Hybrid LLM, the router makes decisions based on query difficulty and desired quality level \citep{ding2024hybrid}.

Self-reinforcing routing optimization utilizes reinforcement learning to progressively improve routing strategies, using the actual performance of the models as feedback signals to continuously optimize and adjust routing policies. MixLLM uses a context-aware contextual-bandit algorithm to learn the optimal query-model allocation, thereby adapting to changing demands and model capabilities \citep{wang2025mixllm}.

Intelligent routing mechanisms can dynamically assign tasks to the most suitable model based on predefined criteria or learned policies. This includes routing based on query difficulty, performance prediction, and cost awareness \citep{wang2025mixllm,mohammadshahi2024routoo}. Smaller, faster models can be used for initial intent detection, routing requests to more specialized or larger models \citep{guha2024smoothie}. Effective task allocation and routing are crucial for optimizing the performance and efficiency of collaborative LLM-SLM systems, ensuring each task is handled by the most appropriate model. Developing robust and adaptive routing mechanisms requires a deep understanding of each model's capabilities and the characteristics of the tasks being processed. This might involve creating benchmarks to evaluate model performance on different task types and designing routing strategies that can adapt to changing conditions.

\subsubsection{Inter-Model Communication and Interface Design}
Seamless and efficient communication between LLMs and SLMs is crucial in collaborative environments.

Natural language interfaces play an important role in this process, particularly in Chain-of-Agents \citep{zhang2024chain}, where working agents can pass information via natural language. This approach leverages the language understanding capabilities of LLMs, simplifies the complexity of interface design, and improves system interpretability \citep{zhang2024chain}.

Structured intermediate representations can provide a more standardized communication mechanism. CoGenesis \citep{zhang2024cogenesis} uses a 'sketch' format for communication between large and small models. This representation conveys key information as a structured summary, making it easier for downstream models to process and more compact yet information-rich than raw text \citep{zhang2024cogenesis}.

Through probability distribution sharing, models can communicate uncertainty at a deeper level. In FS-GEN \citep{zhang2024fast} and CoGenesis \citep{zhang2024fast}, logits sharing allows large models to pass token probability distributions to small models. This not only conveys the final prediction but also the relative likelihood of choices, allowing the small model to make more informed decisions by combining this information with local context \citep{zhang2024cogenesis,zhang2024fast}.

This involves different communication methods, such as passing text-based prompts and responses, sharing intermediate representations (like embeddings, hidden states), or using specialized APIs for inter-model communication \citep{king2024thoughtful,pozdniakov2024large,wozniak2024personalized}. Designing standardized interfaces that allow different models to interact effectively, regardless of their underlying architecture, is also an important consideration. Additionally, challenges related to ensuring data consistency and format compatibility between models need to be addressed. The efficiency and effectiveness of collaboration are directly influenced by the degree of communication and information exchange between models. Well-designed interfaces and communication protocols are essential for achieving seamless interaction. The choice of communication methods and interface design should be tailored to the specific collaboration mode and the type of information being exchanged. Pipeline collaboration might only require passing final outputs, whereas more tightly integrated models may benefit from sharing intermediate representations.

\subsubsection{Model Fusion and Result Integration}
Model fusion refers to merging the architectures or parameters of LLMs and SLMs to create a superior model, including techniques like weight averaging, knowledge fusion, ensemble learning, probability distribution-based fusion, model stacking, and mixture-of-experts models \citep{tang2024small,wan2024knowledge,yang2024model,mavromatis2024pack,wang2025speculate,shi2024profuser}.

Result integration refers to combining the outputs of collaborative models. Common methods include simple averaging, weighted averaging, majority voting, or using another model to combine the outputs. Additionally, handling potential inconsistencies or conflicts in the outputs of different models is necessary. Model fusion and result integration are crucial for creating unified systems that can leverage the distinct strengths and knowledge of multiple models. The choice of technique depends on the specific goals of the collaboration and the characteristics of the involved models. Evaluating the effectiveness of model fusion and result integration requires careful consideration of appropriate metrics and benchmarks to assess the performance of the combined system.

\subsubsection{State Synchronization and Context Management}
In collaborative LLMs and SLMs, especially in multi-turn dialogues or sequential tasks, maintaining a consistent state and managing context is crucial. This involves synchronizing the internal states (memory, attention weights) of the collaborating models and managing and sharing contextual information between them \citep{subramanian2025small,wang2024comprehensive,naveed2023comprehensive,mojarradi2024improving}, for example, using shared memory modules or passing context along with input. Additionally, challenges related to handling long context scenarios and ensuring contextual coherence across multiple interactions need to be addressed.

Maintaining state consistency and effectively managing context are essential for ensuring coherent and meaningful interactions in collaborative LLM-SLM systems, especially for tasks involving multiple steps or turns. The complexity of state synchronization and context management increases with the number of collaborating models and the length of interactions. Developing scalable and efficient techniques to handle these aspects is an ongoing research challenge.

\subsubsection{Dynamic Resource Scheduling and Optimization}
In collaborative LLM-SLM systems, especially when dealing with dynamic workloads and changing resource demands, efficiently allocating and managing computational resources (such as GPU memory, processing time) is a significant challenge \citep{wang2024comprehensive}.

Dynamic resource scheduling techniques can adjust resource allocation based on the current needs of the collaborating models and available resources. Optimization strategies, such as quantization, pruning, and efficient attention mechanisms, can be used to reduce the computational cost and memory footprint of collaborative inference \citep{huang2025dynamic,amayuelas2025self,sun2024llumnix}. The combination of cloud computing and edge computing also offers potential for resource management, for example, LLMs can run in the cloud while SLMs run on edge devices \citep{lv2025collaboration}. Efficient resource scheduling and optimization are essential for making collaborative LLM-SLM systems practical and cost-effective, especially in resource-constrained or high-traffic real-world applications. The heterogeneity of LLMs and SLMs in terms of resource requirements and performance characteristics adds complexity to resource scheduling and optimization. Policies need to consider these differences to achieve optimal resource utilization.

\section{Application Scenarios Driven by On-Device Needs}
With the rapid development of Large Language Models (LLMs), the demand for their application in diverse scenarios is growing. However, the high computational resource requirements of LLMs conflict significantly with the limited computing capabilities of edge devices. Small Language Models (SLMs), due to their lightweight and efficient characteristics, offer a potential solution to this contradiction. This section analyzes typical application scenarios where LLMs and SLMs collaborate, driven by actual on-device needs.
\subsection{Real-time Low-Latency Inference Scenarios}
Real-time low-latency inference is a core requirement of edge computing, especially in applications needing immediate responses, such as wearable/IoT voice assistants and industrial equipment anomaly detection. The collaboration between LLMs and SLMs significantly reduces inference latency and improves user experience and system efficiency through model optimization and edge-cloud collaborative frameworks.

\begin{figure}[t]
 \centering
  \includegraphics[width=0.4\columnwidth]{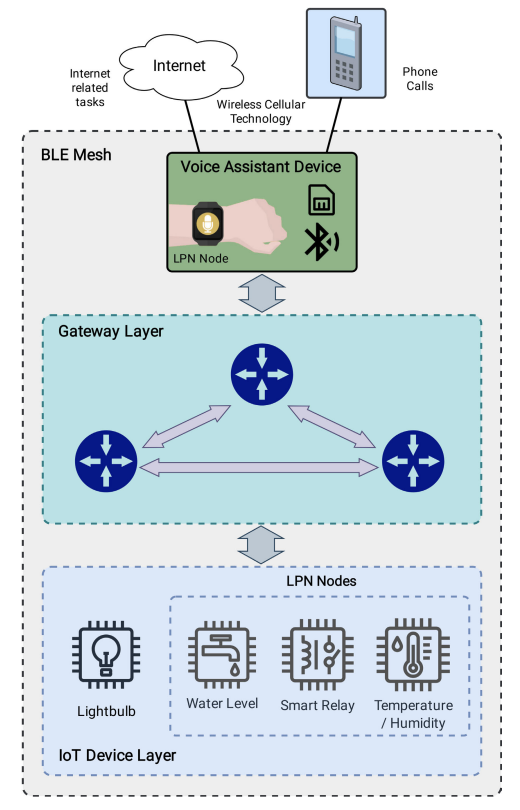} 
  \caption{Edge-cloud collaborative LLM-SLM architecture}
  \label{fig:LLM_SLM}
\end{figure}

Wearable devices and IoT voice assistants require rapid processing of user commands to provide a smooth interactive experience. The powerful language understanding capabilities of LLMs make them ideal for voice assistants, but their high computational demands necessitate optimization through SLMs and edge computing. Research by Froiz-Miguez et al. \citep{froiz2023design} shows that traditional voice assistants relying on cloud processing suffer significant user experience degradation in weak network environments. To address this, researchers have proposed an edge-cloud collaborative LLM-SLM architecture, as shown in Figure~\ref{fig:LLM_SLM}.

\begin{figure}[t]
 \centering
  \includegraphics[width=0.6\columnwidth]{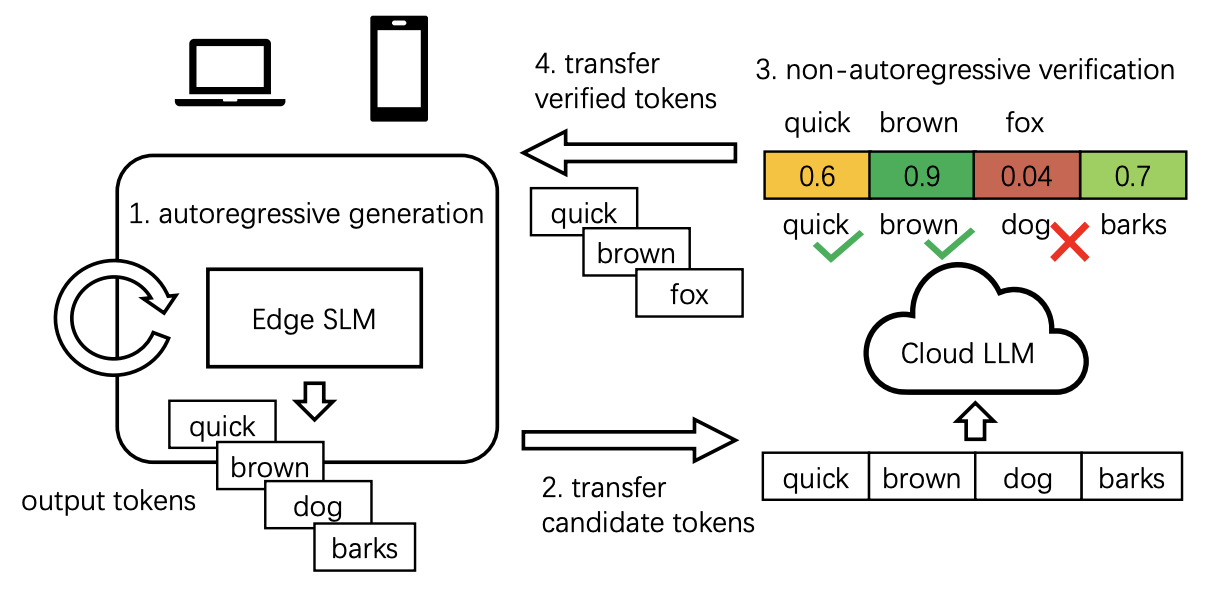} 
  \caption{Hybrid SLM-LLM framework structure}
  \label{fig:Hybrid SLM-LLM}
\end{figure}

The Hybrid SLM-LLM framework proposed by Hao et al. \citep{hao2024hybrid} demonstrates a dynamic token-level edge-cloud collaborative inference method, depicted in Figure~\ref{fig:Hybrid SLM-LLM}. In this approach, a lightweight SLM is deployed on the edge device to handle basic voice commands and generate preliminary responses. Complex tasks are routed to the cloud LLM for further processing based on an uncertainty quantification mechanism. This method achieves an optimal balance between latency and accuracy, particularly suitable for resource-constrained wearable device scenarios. Experiments show that compared to a purely cloud-based mode, this method can reduce end-to-end latency by up to 40\% while maintaining high response quality.

In practical applications, Google's Gboard and SwiftKey have adopted similar methods, integrating local SLMs into mobile keyboards for speech recognition and text prediction, only calling cloud services when more complex understanding is needed \citep{kumar2025large}. This collaborative approach significantly enhances the smoothness of user experience and privacy protection levels.

Device monitoring and anomaly detection in industrial scenarios require systems to respond within milliseconds to prevent potential accidents. Research indicates that while LLMs have strong analytical capabilities, deploying them alone struggles to meet the real-time requirements of industrial settings \citep{zheng2025review}.

To tackle this challenge, researchers have proposed the "Speculative Decoding" mechanism, using edge-deployed SLMs for initial anomaly identification and risk assessment, only activating more comprehensive LLM analysis when potential risks are detected \citep{friha2024llm}. Research by Zheng et al. \citep{zheng2025review} shows that this collaborative mode can reduce response time in industrial anomaly detection from seconds to milliseconds, while maintaining anomaly detection accuracy comparable to pure LLM solutions.

\begin{figure}[t]
 \centering
  \includegraphics[width=0.9\columnwidth]{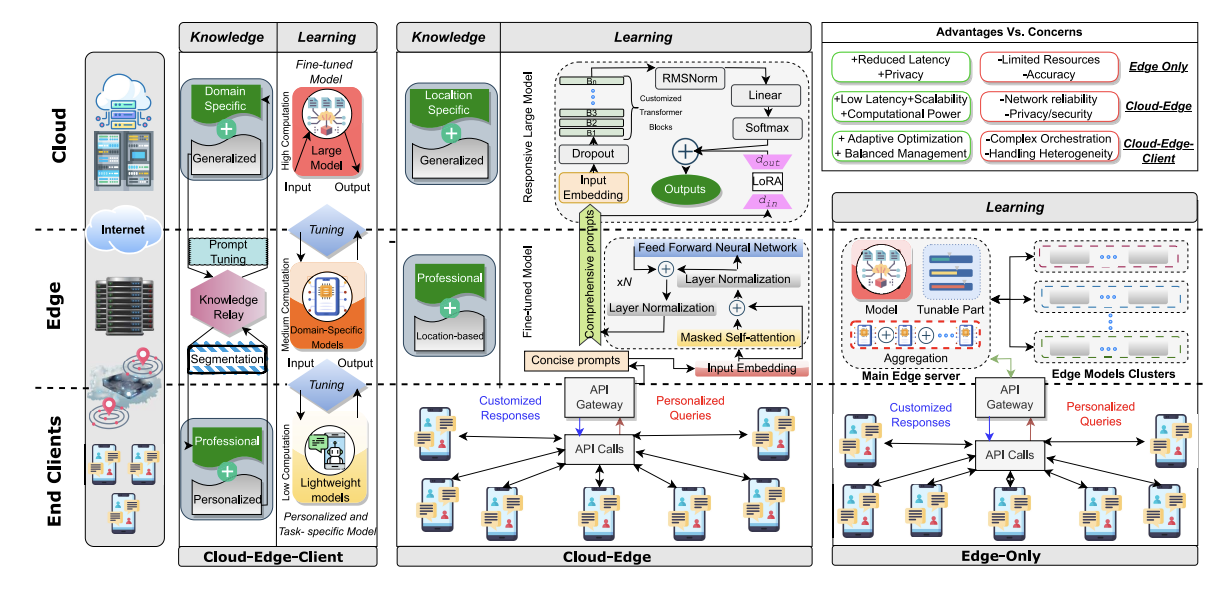} 
  \caption{LLM Edge Deployment Strategy}
  \label{fig:LLM_Edge_Strategy} 
\end{figure}

Models like WinCLIP and AnomalyGPT are specifically optimized for industrial anomaly detection scenarios. Through the collaboration of on-device SLMs and cloud-based LLMs, they achieve dual optimization of accuracy and real-time performance \citep{zheng2025review}. This paradigm demonstrates significant value in high-risk industrial scenarios such as chemical production and power transmission.

\subsection{Privacy-Sensitive and Local Data Scenarios}
In scenarios involving sensitive data, such as personal knowledge base retrieval on mobile terminals and medical image pre-screening, local data processing is key to protecting privacy. The collaboration of LLMs and SLMs meets the needs for privacy protection and data security through local deployment and efficient inference.

Personal knowledge base retrieval and Q\&A systems on mobile devices need to respond quickly to user queries while ensuring data does not leave the device to protect privacy. Personal knowledge bases often contain sensitive information, and uploading them directly to the cloud for LLM processing carries privacy leakage risks. To solve this, researchers proposed privacy-preserving local knowledge base retrieval frameworks. Chen et al. \citep{chen2024llm} pointed out that deploying SLMs on mobile devices can effectively handle local data containing sensitive information without transmitting raw data to the cloud.

A typical implementation is the "Privacy-Sensitive Retrieval-Augmented Generation" (Privacy-Sensitive RAG) architecture. In this architecture, a local SLM is responsible for sensitive data processing and preliminary retrieval, sending only de-identified queries to the cloud LLM for in-depth analysis \citep{wang2024comprehensive}. This method protects user privacy while still fully utilizing the powerful semantic understanding capabilities of LLMs.

FedCoLLM \citep{fan2024fedcollm} further proposed a federated learning framework enabling SLMs on mobile devices to acquire knowledge from cloud LLMs while protecting user privacy data. This framework enhances the capabilities of on-device SLMs through parameter-efficient joint tuning mechanisms, providing a new approach for privacy-preserving knowledge bases on mobile terminals.

Medical data is highly sensitive. The traditional practice of sending data to the cloud for analysis not only poses privacy risks but may also violate relevant regulations. Research shows that deploying SLMs locally can effectively address this issue \citep{niu2025collaborative}.

Zheng et al. \citep{zheng2025review} proposed a "hierarchical diagnosis" model, assigning preliminary screening and basic diagnostic tasks to locally deployed SLMs, only invoking cloud LLMs when encountering complex cases or requiring expert opinions. Experiments demonstrate that this method performs well in diagnosing common diseases while significantly reducing the transmission of sensitive medical data.

Medical domain-specific models like BioMistral \citep{lan2019albert} and PathChat \citep{lu2024multimodal} adopt similar architectures. On-device SLMs process basic medical data and perform initial screening, while cloud LLMs provide more specialized diagnostic advice. This collaborative mode protects patient privacy while also improving the accessibility of medical services, especially in remote areas and resource-limited medical institutions.

\subsection{Task-Specific Customization and Personalization Scenarios}
Through fine-tuning and adapter techniques, LLMs and SLMs can be customized for specific tasks, meeting the personalized needs of individuals and enterprises, such as personal writing assistants and internal corporate knowledge management.

Personalized writing assistants require a deep understanding of the user's long-term writing style and preferences, along with real-time responsiveness \citep{liu2024contemporary}. Traditional cloud-based LLM solutions, while powerful, lack long-term memory of personal data and rapid adaptation capabilities. Researchers proposed a "Hybrid Memory Architecture," maintaining user writing style features, vocabulary preferences, and common expressions in a locally deployed SLM, forming a personalized writing archive \citep{deng2023mutual}. When the user needs deeper creativity or complex content generation, the system calls the cloud LLM for support and feeds the interaction results back to the local SLM to continuously optimize personalized services.

Experiments by Liu et al. \citep{liu2024contemporary} show that this collaborative mode increases the personalization of writing suggestions by up to 35\% compared to pure cloud or pure local solutions, while reducing response time by 70\%. Mainstream document editing tools like Microsoft Word and Google Docs have begun adopting similar architectures, providing instant writing suggestions via local models and delegating complex tasks to the cloud.

Enterprise knowledge management systems need to handle large amounts of specialized domain knowledge while ensuring information security and rapid access. The CE-CoLLM framework proposed by Jin et al. \citep{jin2024collm} designs a cloud-edge collaborative LLM inference architecture for enterprise scenarios, supporting both low-latency edge-independent inference and high-precision cloud-edge collaborative inference modes.

Under this framework, enterprises can deploy SLMs on internal servers to handle common queries and basic tasks, while efficiently invoking external LLM resources when necessary through an "early-exit mechanism" and a "cloud context manager" \citep{jin2024collm}. This approach ensures sensitive information security while meeting high-performance requirements.

The enterprise-grade customized SLM solution developed through collaboration between SoftBank and Aizip further proves the value of this approach \citep{Aizip2024Softbank}. SLMs fine-tuned with domain-specific data are deployed within the enterprise environment, capable of handling unique enterprise tasks with accuracy comparable to cloud LLMs 100 times larger, while significantly reducing latency and costs \citep{Aizip2024Softbank}.

\subsection{Offline or Weak Network Environment Scenarios}
In environments with no network or unstable network connectivity, the local operation of LLMs and SLMs provides crucial support, applicable to scenarios like ocean-going vessels, forest inspection drones, and emergency command at disaster sites.

Ocean-going vessels and forest inspection drones often operate in environments with limited or no network coverage, where traditional cloud-reliant LLMs are ineffective. To address this, researchers proposed an "offline-first" SLM-LLM collaborative architecture. Practice shows that SLMs fine-tuned with domain knowledge can perform most basic interaction tasks completely offline. Research by Kalita \citep{kalita2024large} found that through sufficient fine-tuning on specific domain data, SLMs with 4-5 billion parameters can handle 70-80\% of common queries in offline environments, with performance approaching that of online LLMs. These systems synchronize with cloud LLMs when network connectivity is restored, updating the local model's knowledge base and parameters \citep{kalita2024large}.

A typical application is a drone system for forest inspection. By equipping drones with SLMs fine-tuned with forestry knowledge, they can identify and analyze vegetation conditions, fire risks, and other key information offline, significantly improving inspection efficiency and reliability \citep{lee2025multi}.

In emergency situations like disaster response and border rescue, network infrastructure may be damaged, and the real-time performance of decision support systems directly impacts rescue effectiveness. Research by Tong et al. \citep{tong2024robots} indicates that traditional cloud LLMs lack reliability in such scenarios, while purely local models are often incapable of handling complex situations due to limited capabilities.

To address this, researchers developed a "Resilient Collaboration" framework. Under normal network conditions, the local SLM collaborates with the cloud LLM, forming a complete decision support chain. When the network is interrupted, the local SLM can operate independently using pre-cached knowledge, ensuring basic decision support capabilities \citep{boateng2025survey}.

A successful case is a multi-agent collaborative system used in earthquake rescue. This system integrates offline-deployed small language models and remote large language models, enabling rescue personnel to receive basic decision support even during network outages, and immediately access more comprehensive analysis once the network recovers \citep{Jayant2024StateEdgeAI_misc}. Experiments show that this resilient collaboration solution offers significantly better decision support capabilities in extreme environments compared to single-model solutions.

\subsection{Energy-Constrained and Green AI Scenarios}
In energy-constrained scenarios, SLMs and optimized LLMs meet the needs of edge devices through low-power design, such as keyboard prediction on ultra-low power terminals and driving safety prompts in in-vehicle systems.

On terminal devices with strict power constraints, like smartwatches and limited-functionality IoT devices, the traditional cloud LLM mode significantly reduces battery life due to continuous network connection requirements. Research shows that functions like keyboard prediction, while not requiring LLM-level understanding, demand extremely low latency and energy consumption \citep{qu2025mobile}.

Research by Kumar \citep{kumar2025large} introduced an ultra-low power SLM design specifically optimized for keyboard prediction. Through extreme model compression and hardware-specific optimization, the power consumption for prediction functions was reduced to 1/10th of traditional methods. Although these models have only millions of parameters, they perform well in specific scenarios.

The mobile keyboard prediction SLM developed in collaboration with GloVe further demonstrates the value of SLMs in this scenario \citep{kumar2025large}. The system only wakes up more powerful but energy-intensive models in complex contexts or when explicitly requested by the user, achieving an optimal balance between prediction accuracy and energy consumption, thereby extending device usage time.

In-vehicle systems face unique energy consumption and real-time challenges. Excessive power consumption can affect the vehicle's overall electrical system, while safety-related prompts require millisecond-level responses \citep{lin2023tiny}. Traditional LLM modes struggle to meet these demands.

Qu et al. \citep{qu2025mobile} proposed a "Hierarchical Safety Prompt" architecture. By deploying an SLM specialized in safe driving knowledge within the vehicle system for real-time environment monitoring and basic safety prompts, while retaining connectivity to a cloud LLM for complex situational analysis, this architecture reduces energy consumption by 65\% while maintaining the accuracy of safety prompts.

Several automakers have begun adopting similar "Green AI" solutions, using local SLMs to handle navigation prompts, voice control, and basic safety warnings, only activating more powerful models when deep driving assistance is needed \citep{qu2025mobile}. This approach not only significantly reduces energy consumption but also improves system reliability, especially in areas with unstable network connections.


\section{Challenges and Open Issues}
The collaboration between Large Language Models (LLMs) and Small Language Models (SLMs) aims to combine the strengths of both: the powerful general capabilities and knowledge scope of LLMs, and the efficiency, low latency, and low cost of SLMs on specific tasks. However, achieving perfect collaboration is not straightforward and faces numerous challenges and unresolved open questions. This section delves into these challenges, focusing primarily on collaboration efficiency and system overhead, inter-model consistency and compatibility, robustness and optimality of task allocation strategies, training and maintenance of collaborative systems, evaluation metrics and benchmarking, and related security, privacy, and ethical issues. 
\subsection{Collaboration Efficiency and System Overhead}
\subsubsection{Latency of Task Routing Decisions}
Although LLM-SLM collaboration aims to enhance overall efficiency, the collaboration process itself can introduce new overheads, potentially negatively impacting system performance. Key challenges lie in the latency of task routing decisions, the cost of inter-model communication, and the complexity management of the overall system.

In LLM-SLM collaborative systems, the task routing mechanism is responsible for determining which model should handle a specific query or sub-task, which is core to achieving collaborative efficiency. However, the routing decision process itself introduces latency, potentially partially offsetting the speed advantage gained from using SLMs \citep{arcee2025slms}. The challenge is to design routing mechanisms that are both accurate and fast.

SLMs are favored for their low inference latency, especially suitable for edge devices. In contrast, LLMs, due to their massive parameter count and high computational demands, typically have higher inference latency \citep{wang2024comprehensive}. Collaborative systems require a routing mechanism to dynamically select models based on task complexity or other criteria \citep{arize2025survey}. These routing decisions, whether based on uncertainty estimation, token-level analysis, task decomposition, or other predictive methods, incur their own computational overhead and latency. Cascading methods (querying SLM first, then escalating to LLM if needed) explicitly introduce latency and potential redundancy. Frameworks like CITER attempt to minimize LLM calls through token-level routing but still require a routing mechanism \citep{zheng2025citer}. Pre-generation routing aims to minimize latency by inferring LLM capabilities beforehand, but this prediction itself can be complex \citep{varangot2025doing}.

A core paradox exists: the primary motivation for using SLMs is often to reduce latency, yet the mechanism for effectively choosing when to use an SLM (i.e., the router) itself introduces latency. The router's latency becomes a critical factor; if it is too high, the net latency reduction from collaboration (SLM latency + router latency vs. LLM latency) diminishes. This presents an open research challenge: developing near-zero-latency routing mechanisms or demonstrating that routing overhead is consistently negligible compared to the LLM-SLM latency difference.
**Open Issues:** Develop ultra-low-latency routing mechanisms; quantify the trade-offs between routing complexity/accuracy and latency; design systems capable of masking routing latency through parallel processing.

\subsubsection{Cost of Inter-Model Communication}
In collaborative systems, especially in edge-cloud scenarios (SLM runs locally, LLM in the cloud), data transfer between models (e.g., queries, intermediate results, context, model outputs) incurs communication costs and latency. This is particularly challenging for real-time applications or when handling large amounts of data.

Cloud-edge collaboration is a common pattern to balance privacy/latency (on-device SLM) with capability (cloud LLM). Transmitting user data in real-time to the cloud raises concerns about communication costs, latency, and bandwidth limitations. Frameworks like LSC4Rec aim to minimize real-time data transfer by having the LLM send only a candidate list to the on-device small recommendation model (SRM) for re-ranking \citep{lv2025collaboration}. Similarly, MinionS decomposes tasks in the cloud, sends sub-tasks to local SLMs, and aggregates results in the cloud, aiming to reduce the volume and frequency of cloud interactions \citep{narayan2025minions}. However, communication overhead remains a factor, especially when intermediate results or substantial context need to be exchanged.

While edge-cloud collaboration leverages the efficiency of local SLMs and the power of cloud LLMs, the communication channel itself can become a bottleneck, limiting the overall responsiveness and cost-effectiveness of the system. The design of collaboration protocols and the nature of task decomposition are critical for managing communication overhead. Optimizing communication content and timing is an open challenge.
**Open Issues:** Design communication-efficient collaboration protocols; develop techniques for compressing intermediate representations exchanged between models; quantify the impact of network conditions on collaborative system performance.

\subsection{Inter-Model Consistency and Compatibility}
\subsubsection{Ensuring Consistency in Output Style and Knowledge Scope Across Different Models}
LLMs and SLMs are often trained on different datasets or fine-tuned for different purposes, which can lead them to exhibit distinct writing styles, tones, or response formats. When different models contribute to the same output or dialogue, ensuring a consistent user experience requires mechanisms to align or manage these stylistic differences.

Consistency is considered a fundamental aspect of LLM trustworthiness. Techniques like "Chain of Guidance" (CoG) aim to improve the semantic consistency of LLM outputs through guided prompting \citep{raj2025improving}. Frameworks like SAG explicitly address style alignment in LLM-SLM collaboration using co-training (S-SFT, C-DPO) and self-improvement methods \citep{xusag}. The Modular Pluralism framework uses smaller "community LMs" to integrate diverse perspectives/styles into a base LLM \citep{feng2024modular}. Evaluating consistency itself is challenging, often requiring nuanced metrics beyond simple similarity \citep{raj2025improving}. The KnowsLM framework uses an LLM as a judge to evaluate style adaptation, finding fine-tuning better than RAG for tone adaptation \citep{harbola2025knowslm}.

Collaboration inherently risks stylistic clashes. In the CoGenesis framework, if the SLM handles personal details while the LLM handles the broader narrative \citep{zhang2024cogenesis}, the combined output might feel disjointed unless specific alignment techniques are employed. This necessitates standardizing styles during training/fine-tuning or developing sophisticated output fusion techniques. Since LLMs and SLMs often have different training histories and objectives, this leads to potentially divergent output styles, tones, and formats. In collaborative systems, different models might generate parts of the final output. Combining outputs from stylistically different models can result in an inconsistent and jarring user experience. Ensuring consistency requires explicit effort, either through joint training/fine-tuning for style alignment \citep{xusag}, or through post-processing/fusion techniques \citep{pei2025mathfusion}. This adds complexity to system design and evaluation.
**Open Issues:** Develop robust metrics for evaluating stylistic consistency across different models; create efficient techniques for real-time style alignment or transfer in collaborative generation; understand the trade-offs between stylistic consistency and other objectives like factual accuracy or creativity.

\subsubsection{Alignment of Model Knowledge Scope and Factual Consistency}

Due to differences in training data and model capacity, LLMs and SLMs possess different knowledge bases. LLMs typically have broader knowledge but may hallucinate or contain outdated information; SLMs may have more specialized, potentially up-to-date domain knowledge but lack breadth. Ensuring factual consistency and resolving knowledge conflicts when models collaborate is a significant hurdle.

LLMs suffer from hallucination and knowledge staleness issues, while SLMs often underperform in specialized domains without sufficient domain-specific knowledge. Techniques like Knowledge Distillation (KD) aim to transfer knowledge from LLMs to SLMs \citep{wang2024comprehensive}, but aligning complex distributions is difficult \citep{peng2024enhancing}. RAG is used to inject external, up-to-date knowledge, but consistently integrating retrieved knowledge is also challenging. Research with the KnowsLM framework shows RAG excels at real-time knowledge injection, while fine-tuning is better for stylistic consistency \citep{harbola2025knowslm}. The CoVer framework uses SLM verification to check the consistency of LLM reasoning \citep{yan2025collaborative}. The CrossLM framework uses feedback from SLMs to improve synthetic data generated by LLMs for mutual enhancement \citep{deng2023mutual}.

When LLMs and SLMs possess conflicting information (due to different training cutoffs or specializations), the collaborative system needs a mechanism to resolve this conflict. Simple fusion could lead to factual errors. This requires a sophisticated truth discovery mechanism or a strategy to explicitly trust one model's knowledge domain over another based on the specific query type. LLMs and SLMs have different knowledge bases; LLMs can hallucinate or be outdated, while SLMs may lack broad knowledge or domain specifics. Collaboration can lead to models providing conflicting factual information. The system needs a strategy to handle these conflicts: prioritizing one model based on context/task, using external verification (like RAG), or employing a fusion mechanism that weighs confidence/recency. Failure to resolve conflicts results in unreliable and untrustworthy outputs.

This necessitates research into robust knowledge fusion and conflict resolution mechanisms in LLM-SLM collaborative systems.
**Open Issues:** Develop effective knowledge fusion techniques for LLM-SLM outputs; create dynamic conflict resolution mechanisms based on source reliability or context; ensure traceability of information sources in collaborative outputs; align knowledge updates across collaborating models.


\subsection{Evaluation Metrics and Benchmarking} 
Comprehensively evaluating the performance of LLM-SLM collaborative systems is complex, requiring measurement not only of the quality of the final results but also consideration of multiple dimensions like efficiency and cost. Currently, the field lacks standardized testing platforms and datasets, making comparison between different collaborative strategies and measurement of progress difficult.

Evaluating LLM-SLM collaborative systems cannot rely solely on traditional NLP metrics; it must also comprehensively consider the efficiency gains and cost savings brought by collaboration. This necessitates an evaluation framework capable of capturing the multifaceted performance of collaborative systems.

The challenges of LLMOps extend beyond deploying single models to include data processing, model training, deployment, maintenance, addressing model drift, and ensuring adaptation to changing data and tasks. For collaborative systems, these challenges are even more severe. Practitioners consider model deployment and monitoring phases both important and difficult. Models are often deployed as independent services, with limited adoption of MLOps principles. Reported issues include difficulties in designing infrastructure architectures for production deployment and integration problems with legacy applications. Many models in production are not monitored; key monitoring aspects include inputs, outputs, and decisions. Challenges also include a lack of monitoring practices, the need to create custom monitoring tools, and selecting appropriate metrics \citep{zimelewicz2024ml}. Acxiom, when using LLMs and LangChain for creating audience segmentation systems, faced challenges in debugging complex workflows. Observability achieved through LangSmith optimized token usage and effectively scaled their hybrid model deployment \citep{zenml2025llmops}. Continuous monitoring of LLMs and proactive detection of anomalies and adversarial behavior are crucial for ensuring their integrity. Integrating LLMs into CI/CD pipelines presents concerns related to computational costs, inaccuracies, error handling, bias, and issues related to development, deployment, maintenance, and ethics \citep{pahune2025transitioning}.

The increased number of components in collaborative systems means increased monitoring complexity and more potential failure points. Each model (LLM, SLM), the routing mechanism, and the communication links between them need to be monitored for performance, resource consumption, error rates, and drift. Correlating logs and metrics from different components to diagnose system-wide issues can be highly challenging. Monitoring and debugging tools specifically designed for this distributed, heterogeneous model environment are needed. Collaborative systems have more components (LLM, SLM, router) than single-model systems. Each component is a potential point of failure and needs individual monitoring for health, performance (latency, throughput), resource usage, and output quality. Additionally, the interactions between components need monitoring, such as data flow, API call success rates, and communication latency. When problems arise (e.g., overall response quality degrades), pinpointing which component or interaction is faulty can be complex, requiring sophisticated logging, tracing, and root cause analysis capabilities. This significantly increases deployment and operational complexity.
**Open Issues:** Develop standardized deployment patterns and best practices for LLM-SLM collaborative systems; create comprehensive monitoring frameworks capable of tracking request flows and performance metrics across multiple models; design automated anomaly detection and fault diagnosis tools for collaborative systems; research how to effectively manage and update individual components in distributed model deployments.

\subsection{Security, Privacy, and Ethical Issues}
While offering potential advantages, LLM-SLM collaborative systems may also introduce or amplify risks related to security, privacy, and ethics. The flow of data between different models, the collaborative decision-making process, and the inherent flaws of the models themselves can all be sources of these problems.
\subsubsection{Privacy Risks of Data Flow Between Different Models}
When data, especially data containing sensitive user information, flows between local SLMs and cloud LLMs (or even between different LLMs), the risk of privacy leakage increases.

Using cloud API calls for LLMs is common due to resource constraints, which inherently raises privacy concerns. SLMs are often deployed on edge devices to protect privacy \citep{wang2024comprehensive}. LLMs might leak personally identifiable information (PII) from training data or interactions. Frameworks like CoGenesis aim to logically prevent privacy leaks by having SLMs process private data locally \citep{zhang2024cogenesis}. The CrossLM framework involves SLMs trained on private client data enhancing cloud LLMs, raising questions about knowledge transfer versus data leakage \citep{deng2023mutual}. Federated collaboration is considered a strategy for balancing privacy.

While cloud-edge collaboration distributes data processing, it also creates more nodes for data transfer or processing, potentially increasing the attack surface for privacy breaches if not carefully designed. LLM-SLM collaboration often involves data flow between models, possibly across trust boundaries (e.g., device-to-cloud). LLMs, if handling sensitive user data, pose privacy risks. On-device SLMs are often used to mitigate these risks by keeping sensitive data local. However, the collaboration itself might require sharing some information (intermediate results, context summaries) that could inadvertently leak private details. Ensuring that shared information is sufficiently anonymized or generalized while still being useful for collaboration is a significant challenge. This requires careful design of data flow protocols and potentially the adoption of privacy-enhancing technologies (PETs) within the collaborative framework.
**Open Issues:** Develop privacy-preserving data sharing protocols for LLM-SLM collaboration; formalize privacy guarantees in hybrid systems; investigate the trade-offs between privacy preservation and collaborative performance; audit data flows to identify potential leakage points.

\subsubsection{Potential Bias Amplification or Security Vulnerabilities in Collaborative Systems}
Individual LLMs and SLMs may carry biases from their training data. When these models collaborate, these biases could be amplified, or new, emergent biases could arise due to interaction dynamics. Ensuring fairness and mitigating bias in the combined system is more complex than for individual models.

LLMs can perpetuate and amplify societal biases present in training data. SLMs, if trained on smaller or less diverse datasets, may also exhibit bias. Iterative learning among LLMs can amplify subtle biases \citep{ren2024bias}. The ViLBias framework notes that combining different modalities (text and image) can reveal biases not apparent in single modalities, suggesting model collaboration might have similar effects \citep{raza2024vilbias}. LLM routers themselves can be vulnerable; if an attacker controls routing to direct queries to biased models, it could be exploited to induce biased outcomes \citep{lin2025life}.

Bias can compound in collaborative systems. If an SLM routes a query based on a biased understanding of user demographics, and the selected LLM then generates content with its own biases, the final output could be doubly biased. The interaction itself might also create novel forms of bias not present in the individual models. Both LLMs and SLMs can be biased due to their training data or architecture. In a collaborative system, multiple models contribute to the final output or decision process. If different models have different biases, these can interact in unpredictable ways. A biased output from one model can become biased input for another, potentially amplifying the initial bias (e.g., a biased SLM routes to an LLM which then reinforces the bias). Mitigating bias in such systems requires evaluating not just individual components but the entire collaborative pipeline and its decision logic.
**Open Issues:** Develop methods to detect and measure bias amplification in LLM-SLM collaborative systems; design fair routing and task allocation strategies; create techniques for debiasing outputs from multi-model systems; understand how different collaborative architectures affect bias propagation.

The complexity and distributed nature of LLM-SLM systems might introduce new security vulnerabilities. Attackers could target communication channels, routing mechanisms, or exploit weaknesses in one model to compromise the entire system. LLMs are susceptible to various attacks like prompt injection, jailbreaking, data poisoning, and PII leakage. Research shows LLM routers themselves have lifecycle vulnerabilities, including adversarial attacks and backdoors, with DNN-based routers being particularly vulnerable. If a router is compromised, it could maliciously route queries to inappropriate models or leak data. Interaction points between LLMs and SLMs (e.g., API calls, data exchange) become potential attack surfaces. If an SLM on an edge device is compromised, it could send malicious inputs to the cloud LLM, or vice versa.

Each additional model and communication link in a collaborative system potentially expands the attack surface. Vulnerabilities in the "weakest link" (a less secure SLM or router) in a collaborative system could compromise more powerful LLMs or the integrity/security of the entire system. LLM-SLM systems consist of multiple components (LLMs, SLMs, routers, communication channels). Each component and interface is a potential attack point. Routers could be attacked to force misrouting (sending all queries to an expensive LLM, or to a compromised/ineffective SLM). Communication channels, if insecure, allow data exchanged between models to be intercepted or tampered with. A compromised SLM could feed malicious data to an LLM (or vice versa), potentially leading to harmful outputs or system takeover. Securing a collaborative system requires a holistic approach, considering the security of each component and their interactions, which is more complex than securing a single monolithic model.
**Open Issues:** Develop secure communication protocols for LLM-SLM collaboration; design robust and attack-resistant routing mechanisms; investigate novel attack vectors targeting hybrid model architectures; create comprehensive security auditing frameworks for multi-model systems.

\section{Future Trends}

\subsection{Smarter and More Adaptive Collaboration Frameworks}
The collaboration between Large Language Models (LLMs) and Small Language Models (SLMs) is evolving from preset, static interaction modes towards intelligent frameworks capable of dynamically learning, adapting, and optimizing their collaboration strategies. The core of this trend lies in endowing collaborative systems with greater autonomy, enabling them to adjust flexibly based on task characteristics, data changes, and environmental feedback, rather than relying solely on fixed rules or manual configuration. Future collaborative frameworks will increasingly leverage advanced machine learning techniques, such as reinforcement learning and meta-learning, to achieve dynamic policy adjustments and efficient self-optimization.

\subsubsection{Dynamic Collaboration Strategies Based on Reinforcement Learning}
Reinforcement Learning (RL) provides a powerful paradigm for training agents to make optimal decisions in complex dynamic environments. In LLM-SLM collaboration, RL can be used to learn policies for task allocation, resource management, and interaction protocols to maximize system performance, operational efficiency, or other predefined objectives.

A significant advancement is fine-grained dynamic resource allocation. The CITER framework \citep{zheng2025citer} utilizes RL to train a router that decides at the token level whether the next token should be generated by the SLM or the LLM. The optimization objective of this router considers both prediction quality and inference cost, learning to predict token-level importance and weigh the impact of its decisions on the future generated sequence, thereby achieving a dynamic balance between efficiency and accuracy. This token-level routing strategy is far more granular than traditional query-level routing, allowing the system to leverage SLM efficiency for "simpler" parts of a generation task while relying on LLM power for "critical" tokens, potentially leading to significant efficiency gains.

RL is also being applied to optimize collaboration strategies in specific task scenarios. In Retrieval-Augmented Generation (RAG) tasks, the Collab-RAG framework \citep{xu2025collab} employs an SLM to decompose complex queries and improves the SLM's decomposition ability through feedback signals provided by a black-box LLM. This iterative optimization process, akin to RL mechanisms, enhances the SLM's role in the collaborative RAG process. Furthermore, researchers have proposed the ReMA framework \citep{wan2025rema}, using Multi-Agent Reinforcement Learning (MARL) to guide models towards meta-thinking behavior, where a high-level meta-cognitive agent and a low-level reasoning agent collaborate through RL. This suggests RL can optimize not only task execution but also strategic planning within the collaborative framework. Additionally, Upside-Down Reinforcement Learning (UDRL) has been used to train SLMs for controllable prompt generation to optimize output attributes like length and relevance \citep{lin2025efficient}. While primarily focused on SLM training, the principle of controlling generation attributes via RL can be extended to collaborative scenarios, for instance, when SLM outputs need to meet specific LLM input requirements. Another study proposed a collaborative mode where an LLM generates concise Chain-of-Thought (CoT) instructions, which are then expanded into full responses by an SLM fine-tuned on RL-optimized high-density CoT data \citep{she2025hawkeye}, demonstrating RL's potential in optimizing the role of SLMs within structured collaborative reasoning workflows.

These developments collectively reveal an important trend: RL is driving the shift in collaboration modes from "pre-programmed" to "learning-based." Initial LLM-SLM collaborations often relied on simple, rule-based routing logic. However, as demonstrated by frameworks like CITER, RL can optimize token-level routing decisions based on complex factors like quality and cost, which is far more dynamic and granular than static rules. Works like SmallPlan and Collab-RAG further show that RL can adapt collaboration strategies to specific tasks (e.g., path planning, RAG) and even improve collaborative components (like SLM query decomposers) based on LLM feedback. The ReMA framework pushes RL application to a higher level, enabling hierarchical meta-cognitive and reasoning collaboration. This signifies that RL is not just automating decision processes but teaching the system how to collaborate effectively, driven by data, allowing it to adapt to task nuances and the strengths/weaknesses of different models.

\subsubsection{Dynamic Collaboration Strategies Based on Meta-Learning}
Meta-Learning, or "learning to learn," aims to train models capable of rapidly adapting to new tasks or environments using minimal new data. In LLM-SLM collaboration, meta-learning can be used to develop adaptive routing strategies or collaboration mechanisms that generalize better to different downstream tasks or changing data distributions.

Although explicit applications of meta-learning to LLM-SLM collaboration are still emerging in current research, several directions show strong alignment with meta-learning objectives. Uncertainty-based SLM routing strategies \citep{chuang2025confident}, where queries are passed to a more capable LLM when the SLM is uncertain about its prediction, explore effective generalization of routing policies to new datasets. They propose a data construction pipeline to generate a data-agnostic hold-out set, enabling effective routing decisions without requiring large amounts of new data. This goal of efficient adaptation and generalization is central to meta-learning.

These explorations suggest that meta-learning holds promise for addressing the challenge of rapid adaptation in collaborative frameworks. Collaborative systems need to work effectively across diverse tasks and domains. Retraining complex RL policies for each new task would be costly and inefficient. Meta-learning explicitly trains for rapid adaptation by exposing the learning algorithm to a distribution of tasks during meta-training. Therefore, future research will likely delve deeper into leveraging meta-learning to train the "router" or "coordinator" in LLM-SLM systems, enabling them to quickly configure effective collaboration strategies for unseen new tasks using only a few samples. This would not only significantly reduce the deployment cost and time for LLM-SLM systems in new applications, making them more versatile, but also implies the potential for systems to learn how to learn to collaborate.

\subsection{Deep Fusion of Model Capabilities}
Currently, the collaboration between Large Language Models (LLMs) and Small Language Models (SLMs) is moving beyond simple task routing or pipeline processing towards deeper integration of capabilities. Future trends focus not only on optimizing task allocation but also on exploring interactions between models at finer granularities, potentially even at the neural layer level within the models. The goal is to construct shared understanding or representation spaces, thereby achieving a true fusion of capabilities.

Traditional LLM-SLM collaboration often involves SLMs handling simple tasks or acting as preliminary filters, while LLMs tackle complex tasks. However, future collaboration modes will far exceed this simple hierarchical division, moving towards more complex, bidirectional, and even iterative interaction methods. In such modes, different models might consult each other, verify outputs, or co-construct solutions, forming a dynamic "society of models."

Future collaboration will resemble a "society" of models, where each model plays different roles and follows diverse interaction protocols, rather than a simple hierarchy (e.g., LLM as leader, SLM as helper). Simple routing, while efficient, may not fully leverage the unique strengths of each model type or solve highly complex problems. Introducing mechanisms like LLMs enhancing SLM input data (e.g., SynCID) or SLMs verifying LLM outputs (e.g., CoVer) can create more valuable collaborations. It is foreseeable that more diverse and flexible interaction patterns will emerge, such as iterative refinement, mutual querying, and ensemble methods, where SLMs and LLMs contribute from different perspectives or to different components of a solution. This will lead to more robust and creative problem-solving but also places higher demands on inter-model orchestration and communication mechanisms.

Deeper fusion explores the possibility of LLMs and SLMs interacting at the internal neural network layer level, not just at the input/output level. This could involve sharing activation values, mutual influence on attention mechanisms, or fusing learned intermediate representations. This approach aims to break away from treating each model as an independent black box.

The DeepMLF model \citep{georgiou2025deepmlf}, although designed for multimodal sentiment analysis, offers an inspiring "deep fusion" concept. It introduces learnable tokens that integrate multimodal information at multiple layers of a pre-trained decoder. This concept of cross-layer deep fusion and using learnable tokens as information carriers could be adapted for LLM-SLM collaboration. An SLM could process specific types of features, then inject these features (e.g., activation values) into specific layers of an LLM, and vice versa. This suggests pathways for tighter coupling between models, where one model's internal processing can directly influence another's, potentially yielding richer combined representations. Another study on model pruning \citep{ding2025sliding} proposed a "sliding layer merge" method that dynamically selects and fuses consecutive layers based on a predefined similarity threshold. Although applied to pruning a single LLM, the idea of merging layers based on similarity offers a potential methodology for fusing aligned layers from SLMs and LLMs if their representation spaces can be aligned.

These explorations suggest that the boundaries between LLMs and SLMs might become "softer." Black-box input/output interactions limit the bandwidth and type of information exchangeable between models, as internal representations (activations, hidden states) often contain rich, nuanced information lost in the final output. DeepMLF's success highlights the benefits of deep, multi-layer fusion within a single model, a principle extensible to inter-model collaboration. If an SLM specializes in extracting specific syntactic features, these features (as activations) could be fed directly into relevant layers of an LLM processing the same text for a semantic task, thus enriching the LLM's input at a deeper level. Future research will likely focus on exploring ways to "graft" or "connect" SLMs and LLMs at specific layers, perhaps using adapter modules or learned gating mechanisms to control information flow. This could lead to efficient and powerful hybrid models where SLM strengths (e.g., specialized feature extraction) are deeply integrated into the LLM's broader reasoning process. However, this also presents significant architectural and training challenges, requiring careful alignment of internal representations.


\subsection{Collaboration in Multimodal and Embodied Intelligence}
The collaboration between Large Language Models (LLMs) and Small Language Models (SLMs) is expanding from the purely textual domain into more complex and richer areas of multimodal understanding (integrating vision, speech, etc.) and embodied intelligence (e.g., robots, virtual agents interacting with environments). In these emerging fields, LLM-SLM collaboration will play a crucial role in addressing diverse computational needs and specialized processing tasks.

\subsubsection{Collaboration of Large and Small Models Integrating Multimodal Information like Vision and Speech}
As AI systems increasingly need to process and understand information from multiple modalities, LLM-SLM collaboration becomes essential for effectively managing the different computational demands and specialized processing pipelines for text, images, audio, and other data types. Multimodal data presents unique challenges in terms of data volume, feature extraction, and fusion. LLM-SLM collaboration can efficiently allocate these tasks, with SLMs handling high-throughput, lower-level modality processing, while LLMs focus on high-level cross-modal reasoning and generation.

Existing research envisions LLM agents incorporating "large-small model collaboration" capabilities and handling multi-turn dialogues that may involve multimodal inputs, for example, in search and recommendation scenarios \citep{zhang2025survey}. One study describes training SLMs via LLM distillation to become multitask learners in multimodal contexts, particularly for prompt generation (e.g., text-to-image, text-to-template) \citep{lin2025efficient}, suggesting SLMs can specialize in multimodal-related tasks with LLM assistance. The VITA-1.5 model \citep{fu2025vita} is a Multimodal Large Language Model (MLLM) integrating vision, language, and speech. Although a single MLLM, its multi-stage training approach (progressively integrating visual and speech data) and challenges like modality conflict (speech data potentially degrading visual task performance) highlight the potential benefits of an LLM-SLM collaborative approach. Specialized SLMs could handle initial processing or encoding of specific modalities (like the speech encoder mentioned in VITA-1.5) before feeding features to a central LLM, or manage speech output. The Long-VITA model \citep{shen2025longvita} focuses on long-context vision-language understanding. The complexity and cost of training such models suggest potential roles for SLMs in handling parts of the visual information (e.g., preprocessing, extracting features from segments) or managing shorter-context interactions within a larger mission orchestrated by an LLM.

These trends point towards a "specialized processor" model: In multimodal scenarios, SLMs will likely act as efficient encoders/decoders for specific modalities or as early fusion modules. A vision SLM processes visual input, an audio SLM handles auditory input, and these specialized SLMs feed compact, informative representations to a core LLM that performs cross-modal understanding, reasoning, and generates multimodal responses (perhaps with the help of SLM decoders). This architecture promises more scalable and cost-effective MLLMs, enabling richer human-computer interactions. It also allows for potentially updating or improving modality-specific SLMs independently without retraining the entire MLLM.

\subsubsection{Application of Large-Small Model Collaboration in Robots and Embodied Agents}
Embodied agents (like robots) operate in dynamic real-world environments, requiring real-time perception, decision-making, and action capabilities. LLM-SLM collaboration is well-suited to meet these demands: LLMs provide high-level reasoning and planning, while SLMs handle low-level control, sensor data processing, and rapid reactions on resource-constrained robot hardware.

The SmallPlan framework directly addresses this: An LLM acts as a teacher model to train lightweight SLMs for high-level path planning tasks in robotics. These SLMs, trained via LLM guidance and RL in simulation, can provide action sequences for navigation. This explicitly applies LLM-SLM collaboration to embodied tasks, showing a practical way to distill LLM reasoning into deployable SLMs for robots. One study \citep{glocker2025llm} presents an LLM-powered embodied agent system for autonomous item management by robots in home environments. It integrates multiple specialized agents (routing, task planning, knowledge base) driven by task-specific LLMs (which could include SLMs or LLMs fine-tuned for specialization/smaller size). The system uses RAG for memory augmentation and combines Grounded SAM and LLaMa3.2-Vision for robust object detection, demonstrating how different models (e.g., Qwen2.5 for specialized agents, LLaMA3.1 for routing) collaborate within a complex embodied task. Furthermore, background research on Embodied Multimodal Large Models (EMLMs) \citep{chen2025exploring} highlights the integration of perception, language, and action, pointing out challenges like scalability and generalization where LLM-SLM collaboration could offer solutions (SLMs for onboard perception, LLMs for cloud-based complex planning).

These advancements point towards a hierarchical control and reasoning architecture becoming a dominant paradigm in embodied AI. Robots need both high-level understanding and planning ("clean the kitchen") and low-level, real-time sensorimotor control ("grasp the sponge," "avoid the obstacle"). LLMs excel at commonsense reasoning, task decomposition, and understanding natural language instructions, but they are often too slow or computationally expensive for direct real-time robot control. SLMs, in contrast, are better suited for on-device deployment, handling tasks like local perception, reflexive actions, or executing well-defined sub-goals provided by the LLM. Future systems will likely see LLMs interpreting user goals, breaking them down into actionable steps, and monitoring progress, while a team of SLMs (or a versatile onboard SLM) executes these steps by interacting with the physical world and sending feedback to the LLM. This architecture promises to make robots more intelligent, adaptive, and capable, able to perform complex, long-horizon tasks in human environments, while leveraging powerful cloud-based LLMs alongside onboard SLMs that maintain real-time responsiveness and autonomy.

\section{Conclusion}
The rapid development of Large Language Models (LLMs) has brought significant improvements in artificial intelligence capabilities. However, their high computational resource requirements, deployment costs, and potential latency issues limit their application in broader scenarios. Meanwhile, Small Language Models (SLMs), with their efficiency, low consumption, and ease of deployment, exhibit unique advantages in specific tasks and resource-constrained environments, although their capabilities and generalization are relatively limited. Effectively combining the strengths of LLMs and SLMs to compensate for their respective shortcomings has become a key research topic driving AI technology towards greater efficiency, economy, and universality.

This survey systematically reviews and discusses the collaborative mechanisms and architectures between large and small language models. It first elaborates on the basic concepts, characteristics, and respective advantages and limitations of LLMs and SLMs. Based on this foundation, it focuses on analyzing current mainstream collaboration modes, including pipeline, hybrid/routing, auxiliary/enhancement, knowledge distillation-driven, and integration/fusion collaboration. It also delves into the key technologies for implementing these modes, such as task allocation and intelligent routing, inter-model communication and interface design, model fusion and result integration, state synchronization and context management, and dynamic resource scheduling and optimization. Furthermore, driven by actual on-device needs, this survey explores the practical value of collaborative mechanisms in various application scenarios, including real-time low-latency inference, privacy-sensitive and local data processing, task-specific customization and personalized services, offline or weak network environments, and energy-constrained and Green AI scenarios.

Although research on LLM-SLM collaboration has made significant progress and shown great potential, it still faces numerous challenges. These challenges primarily concentrate on: collaboration efficiency and system overhead, where task routing decisions and inter-model communication may introduce additional latency and costs; inter-model consistency and compatibility, concerning how to ensure uniformity in output style, knowledge scope, and factual accuracy across different models; evaluation metrics and benchmarking for collaborative systems, highlighting the lack of standardized testing platforms and metrics capable of comprehensively measuring collaborative effectiveness, efficiency, and cost; and security, privacy, and ethics, addressing the privacy risks associated with data flow between models, as well as potential bias amplification and security vulnerabilities in collaborative systems. These challenges indicate directions for future research.

\subsection{Research Significance and Value}
Research into the collaborative mechanisms between large and small language models holds profound theoretical significance and broad application value. The core value of collaboration lies in its ability to significantly reduce the demand for computational resources, decrease inference latency, and lower operational costs without substantially sacrificing performance. This enables advanced AI capabilities, previously inaccessible due to high resource barriers, to be deployed on more edge devices and embedded systems, thereby promoting the popularization and democratization of AI technology, benefiting a wider range of users and industries.

As demonstrated in Chapter 4 of this survey, LLM-SLM collaboration can effectively meet the demanding requirements of edge devices regarding real-time response, privacy protection, personalization, offline operation, and low power consumption. Whether it's instant voice assistants on wearables, local knowledge base Q\&A systems protecting personal privacy, or rapid anomaly detection in industrial settings, collaborative mechanisms provide feasible solutions for achieving complex intelligent tasks in resource-constrained environments.

Through reasonable task decomposition and modular design, collaborative systems can assign complex tasks to models with different specialties. This modularity not only helps improve the processing accuracy and efficiency for specific tasks but also makes system debugging, updating, and maintenance more flexible. Specific SLM modules can be targeted for upgrade or replacement without needing to retrain or adjust the entire massive LLM.

With growing concern over AI energy consumption, LLM-SLM collaboration contributes to building more energy-efficient and environmentally friendly "Green AI" systems by optimizing resource utilization and reducing unnecessary computational overhead. This is crucial for the sustainable development of AI technology.

In summary, the collaboration between large and small language models is not only an effective strategy for addressing current AI technology challenges but also a key driving force for stimulating AI application innovation and expanding the boundaries of AI capabilities.

\subsection{Reflections and Outlook}
Looking ahead, research on the collaborative mechanisms between large and small language models will continue to deepen towards directions that are smarter, more integrated, and more broadly applicable.

Many current collaboration strategies still rely on preset rules or relatively fixed task allocations. In the future, dynamic collaboration strategies based on advanced techniques like reinforcement learning and meta-learning will become mainstream. Systems will be able to autonomously learn and optimize collaboration methods based on real-time task complexity, data characteristics, available resources, and even user feedback, achieving self-configuration and self-adaptation. Examples include the token-level dynamic routing in the CITER framework and optimizing the role of SLMs in RAG tasks through reinforcement learning. This will make collaborative systems more flexible and efficient, better able to cope with complex and changing real-world application environments.

Future collaboration will move beyond task-level allocation or simple output combination to explore deeper interactions and capability fusion between models. This might involve information exchange and influence at the internal neural layer level, such as sharing activation values, mutual guidance of attention mechanisms, or even constructing unified or aligned representation spaces. Inspired by techniques like the DeepMLF model and sliding layer merging, breaking the model black box and achieving deep coupling of internal states promises to yield hybrid intelligent agents with stronger capabilities and more seamless coordination.

As AI technology evolves towards multimodal understanding and embodied intelligence, LLM-SLM collaboration will play a central role in processing and integrating diverse information streams, including text, vision, speech, and even physical interactions. In robotics applications, LLMs can handle high-level task planning and understanding, while a series of specialized SLMs on the robot hardware manage real-time sensor data, execute low-level control commands, and provide rapid environmental responses, as demonstrated by the SmallPlan framework. This hierarchical collaboration will be key to building intelligent agents capable of autonomously operating in complex physical worlds.

With the increasing capabilities and widespread application of collaborative systems, issues concerning the transparency of their decision-making processes, potential security risks (such as adversarial attacks and privacy leakage), and possible biases will become more prominent. Future research needs to invest more effort in developing effective interpretability methods, constructing robust security defense systems, and exploring ethical guidelines and governance frameworks suitable for collaborative intelligent systems.

In conclusion, the collaborative mechanism between large and small language models is in a phase of rapid development. It not only provides innovative solutions to the bottlenecks of existing AI technologies but also opens up a new continent full of possibilities for the future of artificial intelligence. We have reason to believe that as theoretical research deepens and application scenarios continue to expand, this collaborative intelligence paradigm will inevitably play an increasingly important role in the future AI ecosystem, driving AI technology towards greater efficiency, universality, intelligence, and responsibility.

\printbibliography  

\end{document}